\documentclass{article}

\usepackage{arxiv}

\usepackage[utf8]{inputenc} 
\usepackage[T1]{fontenc}    
\usepackage[pdftex]{graphicx}
\usepackage{hyperref}       
\usepackage{url}            
\usepackage{booktabs}       
\usepackage{amsfonts}       
\usepackage{nicefrac}       
\usepackage{microtype}      
\usepackage{lipsum}
\usepackage{fancyhdr}       
\usepackage{graphicx}       
\usepackage{caption}
\usepackage{subcaption}
\usepackage{here}
\usepackage{float}
\usepackage{threeparttable}
\usepackage{xcolor}
\graphicspath{{media/}}     

\title{Multi-Scale Spatial-Temporal Self-Attention Graph Convolutional Networks for Skeleton-based Action Recognition
}

\author{
  Ikuo Nakamura \\
  Sony Corporation \\
	\texttt{spin.funnel@gmail.com} \\
}

\begin{document}
\maketitle

\begin{abstract}
Skeleton-based gesture recognition methods have achieved high success using Graph Convolutional Network (GCN).
In addition, context-dependent adaptive topology as a neighborhood vertex information and attention mechanism leverages a model to better represent actions.
In this paper, we propose self-attention GCN hybrid model, Multi-Scale Spatial-Temporal self-attention (MSST)-GCN to effectively improve modeling ability to achieve state-of-the-art results on several datasets.
We utilize spatial self-attention module with adaptive topology to understand intra-frame interactions within a frame among different body parts, and temporal self-attention module to examine correlations between frames of a node.
These two are followed by multi-scale convolution network with dilations, which not only captures the long-range temporal dependencies of joints but also the long-range spatial dependencies (i.e., long-distance dependencies) of node temporal behaviors.
They are combined into high-level spatial-temporal representations and output the predicted action with the softmax classifier.
\end{abstract}


\section{Introduction}
The skeleton-based representations have become very popular for human action recognition because abstracted human skeleton data is compact to depict dynamic changes in human body movements.
Moreover, compared to conventional RGB-based action recognition methods, they are robust to the variations of illumination, colorization, and other background noises.
As a result, it is also a promising solution for various real-world applications.

The dominant solution in skeleton-based action recognition is a graph convolution network (GCN) \cite{gcn1}.
Yan et al.\cite{stgcn} first treat joints and their physical connections as nodes and edges of a graph and employ GCN on such a predefined graph to learn joint interactions.

Eventually, the limitations of expression through physical connections are identified.
In the real world, humans recognize actions relationships between spatially or temporarily distant joints, as well as between adjacent joints are strongly correlated.
Subsequent studies use a learnable topology which uses the physical connections for initialization 
representing dependencies based on prior knowledge and learns spatiotemporal features heuristically from skeleton data.

GCN-based model uses spatial and temporal modules independently of each other, and physically adjacent graph still hinders the model’s ability to capture effective spatial relations of joints across a distant node topology.

On the other hand, transformer-based approach using self-attentions to skeleton-based action recognition has been addressed.
It can learn the global dependencies between the input elements with less computational complexity and better parallelizability. 
Self-attention methodologies increase the spatial receptive field. But they still rely on using physically adjacent graphs, 
which can lead to biased results towards those physically adjacent graphs and highlight a potential limitation in their effectiveness.
To handle this problem, Liu et al. \cite{multiscale} propose a multi-scale graph that identifies the relationship between structurally distant nodes.

Chi et al. \cite{infogcn} propose a hybrid model which incorporates self-attention mechanism to GCN to capture context-dependent intrinsic topology to better representation of actions.
However, it cannot capture the movement of joint along the frames which can be also correlated with those of remote nodes.
To solve this limitation, we propose modules of Temporal Self-Attention (TSA) followed by Multi-Scale Spatial Convolutions to extract temporal features from distant nodes.
We revisit the two-streams architecture of ST-TR \cite{sttr} with refining adaptive spatial-temporal modules.

TSA, that is vanilla self-attention module, provides the intrinsic nature of all the frames for each joint.
The contextual information can be extracted to recognize human action by comparing the change in the embeddings of the same joint along the temporal dimension.
But it lacks inductive bias to capture the locality of spatial structural data. We compensate it by use of following multi-scale spatial convolution \cite{multiscale, mstgnn}.
To verify the effectiveness of our model, extensive experiments are conducted on four skeleton-based action recognition benchmark datasets: SHREC'17 \cite{shrec17}, NTU-RGB+D 60 \cite{ntu60}, and Northwestern-UCLA \cite{nwucla}.

\section{Related Works}
To further capture the interaction between physically unconnected joints,
the attention mechanism is one of the essential elements for modeling such intrinsic topology. 
The model can relax the restriction of fixed topology and emphasize important information along a specific dimension through attentions. 
These techniques are divided into two categories for GCNs:
 (1) attention-based graph construction \cite{2sagcn,ctrgcn}
which is a method of forming topologies using a nonlocal block or customized correlation matrices, and (2)
spatial-wise, temporal-wise, and channel-wise attention, which
are commonly used attentions in \cite{msaagcn, effigcn}. 

Various GCN-based extensions are proposed.
HD-GCN \cite{hdgcn} applies hierarchically decomposed graphs and their attentions to extract major structurally adjacent and distant feature.
STC-Net \cite{stcnet} propose dilated kernel for Graph convolution and Spatial Curve Module which aggregates different nodes for different frames. 
This model takes account of both coordinate and motion data in the multi-modal like manner at the outset.

Several recent works extend Transformer to build attentions.
DSTA-Net \cite{dstanet} is the first to introduce self-attention which forms unified framework based solely on decoupled spatial-temporal self-attention mechanism.
ST-TR \cite{sttr} models both spatial and temporal dependencies of the skeleton sequence with self-attentions.
Zhang et al. \cite{stst} propose a Spatial-Temporal Specialized Transformer (STST), which designs a Directional Temporal Transformer Block for modeling skeleton sequences in temporal
dimensions.

Recently, more structured self-attentions have been proposed considering the intrinsic nature of skeleton.
HyperFormer \cite{hyperformer} introduced the concept of hypergraph to self-attention which is aware of extra higher-order relations of body parts.
STEP CATFormer \cite{catformer} combines CTR-GCN \cite{ctrgcn} with two joint cross-attention transformers, which are for the upper-lower/hand-foot body parts. 
FGSTFormer \cite{fgstformer} propose transformers to capture the discriminative correlations of focal joints as well as global parts in both spatial and temporal dimensions.

A new approach combining knowledge-based data augmentation is also reported. LA-GCN\cite{lagcn} propose a GCN using large-scale language models (LLM) knowledge assistance. Despite the good results in experiments, the attempts of LA-GCN to acquire and apply LLM prior knowledge information are more oriented towards manual design.

Our model uses self-attention based GCN but does not consider any intrinsic nature of skeleton other than physical connections to avoid inclination towards manual design with prior knowledge. 
We still demonstrate the state-of-the-art performance.

\section{Methods}
The human skeleton can be represented as a graph $G(V, E)$ with joint as vertices $V$ and bone as edges $E$, $V = \{v_1, v_2, \cdots, v_N\}$ is the set of $N$ joints.
Edges can be represented as an adjacency matrix $A\in \mathbb{R}^{N \times N}$, where $A_{i,j} = 1$ if joints $i$ and $j$ are physically connected, and otherwise $0$.

Our MSST addresses normalized adaptive-adjacency matrix represented as $\tilde{A}$.
For a skeleton sequence graph of $T$ frames and $N$ joints, let the input of $C$-dimension be $\mathbf{X}\in \mathbb{R}^{T\times N \times C}$.

Firstly, the embedding block linealy transform $\mathbf{X}$ to hidden dimension $D^{(0)}$ as $\mathbf{H}^{(0)}\in \mathbb{R}^{T\times N \times D^{(0)}}$.
Hidden representation of $l$-th layer $H^{(l)}$ can be rearranged as $\mathbf{H}^{(l)}=\{\mathbf{H}^{(l)}_t \in \mathbb{R}^{N\times C}\}_{t=1}^T$
or $\mathbf{H}^{(l)}=\{\mathbf{H}^{(l)}_n \in \mathbb{R}^{T\times C}\}_{n=1}^N$ to tackle temporal dimensions.
The following encoding block deal with the projected streams in two pathway manner, such that we define 1$st$ stream which uses $\mathbf{H}^{(0)}_t$, and 2$nd$ stream for $\mathbf{H}^{(0)}_n$.

\subsection{Self-attentions}
Self-attentions are permutation invariant for the inputs. Hence, at successive layer, 1$st$ stream adapts position encoding (PE) to inject positional information to Query, Key of joints in 
Spatial Self-Attention (SSA) as $SPE\in \mathbb{R}^{N\times D^{(0)}}$.
The 2$nd$ stream injects temporal information of frames in TSA as $TPE\in \mathbb{R}^{T\times D^{(0)}}$.

We utilize vanilla self-attention for SSA which joints for a frame to infer intrinsic topology.
Multi-head attention is applied by repeating extraction process $M$ times each with a different set of learnable parameters each time. 
Then we fuse the outputs of $M$ individual attention components.

We linearly project joint representation $H_t \in \mathbb{R}^{N\times D}$ to queries and keys of $D^{'}$ dimesions with learned matrices $\mathbf{W}^{sp,m}_Q, \mathbf{W}^{sp,m}_K, \mathbf{W}^{sp,m}_V \in \mathbb{R}^{D \times D^{'}}$
for a head $1\le m\le M$ where subscripts $Q, K, V$ represent Query, Key, Value. The letter $sp$ in learned matrices represents $spatial$ of parameters.
Adaptive topology $\tilde{A}$ is combined with self-attention map \cite{2sagcn, infogcn} for each head $m$.
We name it SSA Graph Convolution (SSA-GC) with $m$ head $SSAGC_m(H_t)\in \mathbb{R}^{T\times N\times C}$ for a frame is formulated as 
\begin{equation}
  \label{ssa}
  SSAGC_m(\mathbf{H}_t) = \{\tilde{A}_m\odot softmax(\frac{\mathbf{H}_t \mathbf{W}^{sp,m}_K (\mathbf{H}_t \mathbf{W}^{sp,m}_Q)^T}{\sqrt{D^{'}}})\}\mathbf{H}_t \mathbf{W}^{sp,m}_V
\end{equation}
 
Similarily, TSA with $m$ head $TSA_m(H_n)\in \mathbb{R}^{N\times T\times C}$ for a joint is also defined.
We provide frame representation $H_n \in \mathbb{R}^{T\times D^{'}}$ with learned matrices $\mathbf{W}^{te,m}_Q, \mathbf{W}^{te,m}_K, \mathbf{W}^{te,m}_V \in \mathbb{R}^{D \times D^{'}}$.
The letter $te$ represents $temporal$ of parameters.
\begin{equation}
  TSA_m(\mathbf{H}_n) = softmax(\frac{\mathbf{H}_n \mathbf{W}^{te,m}_K (\mathbf{H}_n \mathbf{W}^{te,m}_Q)^T}{\sqrt{D^{'}}})\mathbf{H}_n \mathbf{W}^{te,m}_V
\end{equation}

In the temporal dimension, it explicitly learns dynamic temporal relations with a dilation convolution transformer, enabling the network to capture rich motion patterns effectively.

\subsection{Multi-Scale Convolution}
To model the temporal feature of the skeleton, we adopt a Multi-Scale Temporal Convolution (MS-TC) \cite{googlenet,infogcn,tdgcn}.
The module consists of four branches, kernel size $5 \times 1$ with dilation rates $d_t=[1,2]$, a maxpooling branch $3 \times 1$ and a residual connection.
The output of different branches is aggregated by concatenation.
In the 1$st$ stream, we combine SSA-GC and MS-TC to capture the long-range temporal dependencies of extracted all feature.

For the spatial feature of the skeleton, we introduce Multi-Scale Spatial Convolution (MS-SC) which consist of the same four branches as MS-TC
except the direction of convolution where the kernel size $1 \times 5$, dilation rates $d_s=[1,2]$ and a maxpooling branch $1 \times 3$. 
In the 2$nd$ stream, TSA followed by MS-SC can extract distant spatial dependencies of dynamic feature of a joint.

\begin{figure}[H]
  \vspace{3cm}
  \begin{minipage}[b]{0.48\columnwidth}
    \includegraphics[bb=0 50 60 200, scale=0.3]{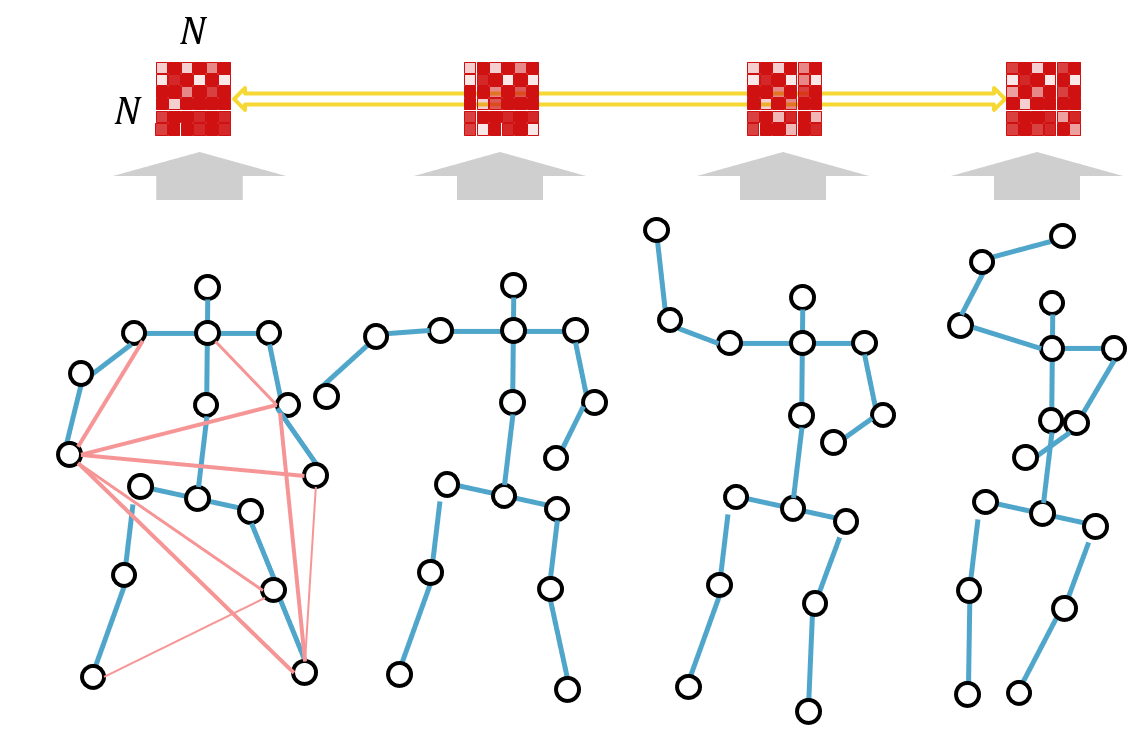}
    \vspace{0.5cm}
    \subcaption{SSA-GC + MS-TC}
  \end{minipage}
  \hspace{0.04\columnwidth} 
  \begin{minipage}[b]{0.48\columnwidth}
    \includegraphics[bb=0 50 60 200, scale=0.3]{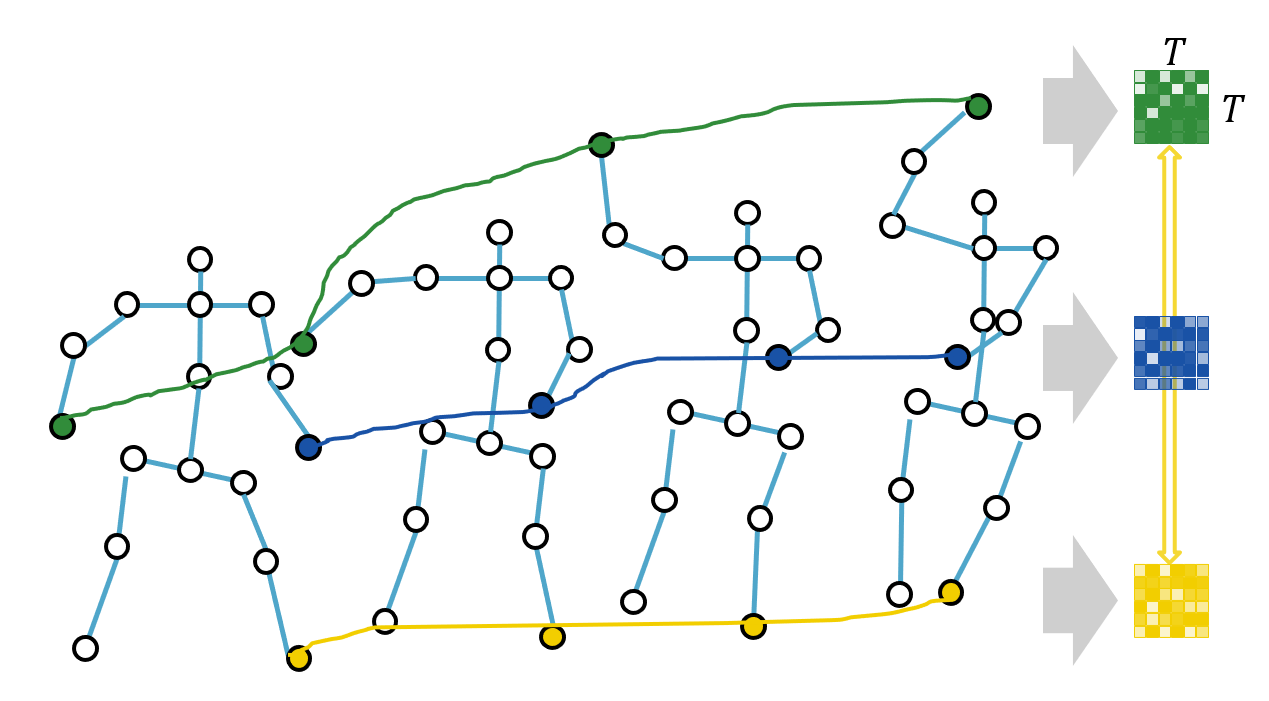}
    \vspace{0.5cm}
    \subcaption{TSA + MS-SC}
  \end{minipage}
  \caption{(a) Spatial Self-Attention operates on each pair of nodes. The self-attention maps of size $N \times N$ are projected to $N \times C$ by Value. 
  Multi-Scale convolution is applied to temporal dimension to extract the long-range temporal dependencies.
  (b)Temporal Self-Attention operates on each pair of frames. The self-attention maps of size $T \times T$ are projected to $T \times C$ by Value.
  Multi-Scale convolution is applied to node dimension to extract the distant spatial dependencies.}
  \label{figure1}
\end{figure}

\subsection{Two stream networks}
Our model is composed of an embedding block and a stack of L = 9 encoding blocks followed by a global average pooling layer (GAP).
By processing SSA-GC followed by MS-TC with a residual, a layer normalization and GAP,  we obtain the first latent representation $\mathbf{z_1}\in \mathbb{R}^{C_{out}}$.
Similarly, TSA followed by MS-SC and GAP, we obtain the second latent representation $\mathbf{z_2}\in \mathbb{R}^{C_{out}}$.
We define a fusion function $F$ which aggregates and updates high-level spatial-temporal feature from the two streams respectively. 
\begin{equation}
  F(\mathbf{z_1}, \mathbf{z_2})=\mathbf{z_1} + \mathbf{\alpha}^T \mathbf{z_2}
\end{equation}
where $\mathbb{\alpha}$ is a learnable parameter optimizing the channel-wise composition rate during the training process.

\subsection{Multi-Scale Spatial-Temporal self-attention GCN}
\begin{figure}[H]
  \vspace{1cm}
  \includegraphics[bb=0 50 100 500, scale=0.4]{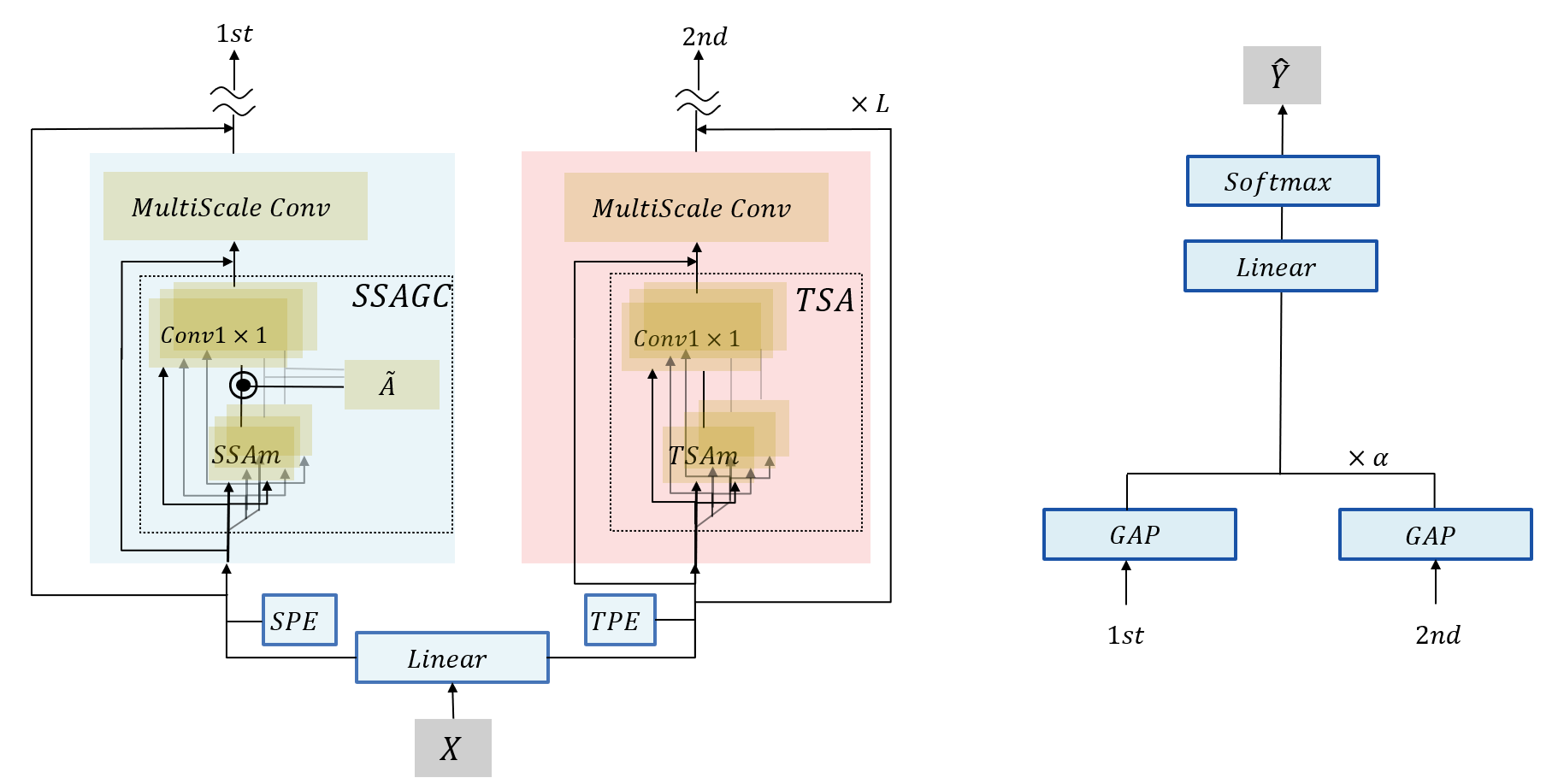}
  \vspace{0.5cm}
  \caption{The overall framework of our MSST-GCN, where $SSAm\in \mathbb{R}^{N \times N}$ represents spatial self-attention map in SSA and $TSAm \in \mathbb{R}^{T \times T}$ represents temporal self-attention map in TSA.
  Three multi-head attentions are described in the figure.}
  \label{figure2}
\end{figure}

The overall blocks of our MSST-GCN is illustrated in Figure \ref{figure2}. After channel-wise composition with parameter $\alpha$, the fusin function with a auxiliary random noise $\epsilon \sim \mathcal{N}(0, I)$ is fed into a fully-connected layer and softmax classifier for predicting action.
The model is updated to minimize the total loss, which we adopt the summation of cross-entropy, marginal Maximum Mean Discrepancy (MMD) and conditional-marginal MMD which are introduced in \cite{infogcn}.

\subsection{Ensemble with Multi-modal representation}
Many state-of-the-art methods use a multi-stream fusion strategy to improve action recognition performance.
Generalized joint-bone relation of inputs at time $t$ is shown as \cite{2sagcn,infogcn}
\begin{equation}
  \tilde{X_t} = (I-P^k)X_t
\end{equation}
where $P \in \mathbb{R}^{N\times N} $ denotes a binary matrix that contains source-target relations of skeleton graph, $P_{ij}=1$ if the $i$-th joint is the source of the $j$-th
joint, otherwise 0.
It corresponds to bone representation if $k=1$, and joint representation if $k=K$ with $P^K = 0$. 

The final results of the experiments fuse four streams (4s), namely, joint position (J), bone position (B), the differentiation of the joint in the time dimension: joint motion (JM), 
and that of the bone: bone motion (BM) respectively.
We also report the results of 6 ensemble (6s) fusing additional $k=2$ position and motion streams.  
After training a model for stream, the softmax scores of each stream are summed up as the final score during inference.

\section{Experiment}
\subsection{Datasets}
SHREC’17 \cite{shrec17} Track dataset contains 2800 gesture sequences performed in two ways: using one finger and the whole hand.
And each gesture is performed between 1 and 10 times by 28 participants. Training data comes from 1960 sequences, 
and the remaining 840 sequences are used for testing. The gesture sequence can be labeled according to 14 or 28 classes, depending on the number of fingers used and the gesture represented.
The recognition accuracy of the SHREC dataset will also be computed following 14 or 28 gesture classes. 
The hand skeleton of the SHREC’17 Track dataset contains 22 joints.

NTU-RGB+D 60. NTU-RGB+D 60 \cite{ntu60} is a large dataset used in skeletal action recognition. 
It contains 56,880 skeleton action samples, performed by 40 different participants and classified into 60 classes. 
The authors of this dataset recommend two benchmarks. (1) Cross-Subject (X-Sub): 20 of the 40 subjects’ actions are used for training,
and the remaining 20 are for validation. (2) Cross-View (X-View): Two of the three camera-views are used for training,
and the other one is used for validation.

Northwestern-UCLA. The Northwestern-UCLA skeleton dataset \cite{nwucla} contains ten basic human actions and 1494 video clips from 10 actors. 
All data were obtained from three Kinect cameras captured simultaneously. 
Following the evaluation protocol introduced in \cite{nwucla}, the data from the first two cameras are used as the training set, and the data from the last camera as the test set.

\subsection{Experimental settings}
We adopted InfoGCN \cite{infogcn} as a backbone.
Sequences of the skeletons are batched together after being resized to 64 frames.
The SGD optimizer is employed with a Nesterov momentum of 0.9 and a weight decay of 0.0004.
The number of learning epochs is set to 120, with a warm-up strategy applied to the first five
epochs for more stable learning. The batch size for all datasets is 64, the learning rate
to decay with cosine annealing, with a maximum learning rate of 0.1 and a minimum learning rate of 0.0001.

In terms of our experiments with 28 gestrues of SHREC'17, NTU-RGB+D 60 and NW-UCLA datasets, we set $base\_channel$ to 64 which makes output channels of 9 layers 64,64,64, 128,128,128, 256,256,256.
For 14 gestures of SHREC'17 dataset , we apply $base\_channel$ to 32 which takes into account of the dataset size and makes output channels as 32,32,32,64,64,64,128,128,128.
We use generalized bone representation $K=8$ for NTU-RGB+D, $K=5$ for SHREC'17 and $K=6$ for NW-UCLA.

\subsection{Experimental results}
We present a comprehensive comparison for our model against recent state-of-the-arts methods in Table \ref{table1}, \ref{table2}, \ref{table3}.
In terms of SHREC'17 dataset, our model attains top accuracy for 14 gestures but underperforms for 28 gestures using four ensembles.
In NTU-RGB+D 60 dataset, it outperforms InfoGCN \cite{infogcn} and obtain comparable result with the state-of-the-arts models.
Our model with 6 ensembles on NW-UCLA dataset outperform any models in the table \ref{table3}.
The model difference among these three datasets are only the number of nodes and physical bone conections, which means our model is universally applicable to any joint models.
 
\begin{table}[H]
  \caption{Classification accuracy comparison on SHREC'17 dataset}
   \centering
   \begin{tabular}{lll}
     \toprule
     \cmidrule(r){1-3}
     Method     &14 Gestures(\%)      & 28 Gestures(\%) \\
     \hline \hline
     DSTA-Net \cite{dstanet}  & 97.0  & 93.9 \\
     MS-ISTGCN \cite{istgcn} & 96.7 & 94.9 \\
     TD-GCN \cite{tdgcn} & 97.0 & 95.4 \\
     \hline
     MSST-GCN 4s &97.0 &94.8 \\
     \bottomrule
   \end{tabular}
   \label{table1}
\end{table}
 
\begin{table}[H]
 \centering
 \begin{threeparttable}
 \caption{Top-1 classification accuracy comparison on NTU-RGB+D 60 dataset. The numbers in gray indicate the result that we retrain the models using their officially released code.}
   \begin{tabular}{lll}
     \toprule
     \cmidrule(r){1-3}
     Method     &X-Sub(\%)      &X-View(\%) \\
     \hline \hline
     GCN-based methods \\
     \hline
     ST-GCN \cite{stgcn} & 81.5  & 88.3     \\
     2s-AGCN \cite{2sagcn} &88.5 &95.1 \\
     CTR-GCN \cite{ctrgcn} &92.4 &96.8 \\
     TD-GCN \cite{tdgcn}  & 92.8 & 96.8 \\
     STC-Net \cite{stcnet}\tnote{a} &93.0 &97.0 \\
     HD-GCN 4s \cite{hdgcn} &93.0 &97.0 \\
     DG-STGCN \cite{dgstgcn}\tnote{b} &93.2 &97.5 \\
     \hline
     Transformer-based methods \\
     \hline
     STTR \cite{sttr} &90.3 &96.3 \\
     DSTA-Net \cite{dstanet}  & 91.5 & 96.4 \\
     STST \cite{stst}  &91.9 &96.8 \\
     FGST-Former \cite{fgstformer} &92.6 &96.7 \\
     HyperFormer \cite{hyperformer} &92.9 & 96.5 \\
     STEP CATFormer \cite{catformer} &93.0 & 96.9 \\
     HD-GCN 6s \cite{hdgcn} &93.4 &97.2 \\
     \hline
     Hybrid model \\
     \hline
     Efficient-GCN \cite{effigcn}&92.1 &96.1 \\
     Info-GCN 4s\cite{infogcn}&92.7 &96.9 \\
     Info-GCN 6s\cite{infogcn}&93.0 &97.0 \\
     \textcolor{gray}{Info-GCN 4s} &\textcolor{gray}{92.3} &\textcolor{gray}{96.6} \\
     \hline
     MSST-GCN J+B &92.3 &96.6 \\
     MSST-GCN 4s &92.8 &97.0 \\
     MSST-GCN 6s &93.0 &97.3 \\
     \bottomrule
   \end{tabular}
   \label{table2}
  \begin{tablenotes}
  \item[a] It uses both position and motion in one model, and different ensemble methods.
  \item[b] Sampling method is introduced for data augmentation. It divides the graph into several feature groups of which number is a hyper-parameter.
  \end{tablenotes}
 \end{threeparttable}
\end{table}

\begin{table}[H]
  \caption{Top1 classification accuracy comparison on NW-UCLA set}
   \centering
   \begin{tabular}{lll}
     \toprule
     \cmidrule(r){1-2}
     Method     &NW-UCLA(\%) \\
     \hline \hline
     CTR-GCN\cite{ctrgcn}  & 96.5  \\
     Info-GCN 4s\cite{infogcn}  & 96.6 \\
     Info-GCN 6s\cite{infogcn}  & 97.0 \\
     HyperFormer \cite{hyperformer} & 96.7 \\
     FGST-Former \cite{fgstformer}  &97.0 \\
     HD-GCN 6s\cite{hdgcn} &97.2 \\
     STC-Net\cite{stcnet} &97.2    \\
     TD-GCN\cite{tdgcn} &97.4    \\
     \hline
     MSST-GCN 4s  &97.8   \\
     MSST-GCN 6s  &98.1   \\
     \bottomrule
   \end{tabular}
   \label{table3}
\end{table}

\subsection{Ablation studies}
Ablation studies are conducted on NTU-RGB+D 60 and NW-UCLA dataset.
We first validate the effectiveness of TSA MS-SC module setting.
\subsubsection*{Influence of TSA MS-SC stream}
In table \ref{comparison}, we have a detailed comparison under the four data modalities with the baseline implementation of InfoGCN \cite{infogcn} model which is retrained using their officially released code.
We also refer to the retraining results described in \cite{skeletongcl}.
We outperform InfoGCN \cite{infogcn} by 0.2\% to 0.6\% in NTU-RGB+D dataset, and by 1.0\% to 2.0\% in NW-UCLA dataset.
\begin{table}[H]
  \caption{\textbf{Comparison of InfoGCN and Ours}}
  \label{comparison}
   \centering
  \begin{minipage}{0.49\textwidth}
    \centering
    \caption*{NTU-RGB+D 60 dataset}
    \begin{tabular}{lll}
     \toprule
     \cmidrule(r){1-3}
     Modality/Method     &X-sub(\%)      & X-view(\%) \\
     \hline \hline
     Joint (InfoGCN) &88.75 &95.25\\
     Joint (Ours) &90.19 &95.43\\
     \hline
     Bone (InfoGCN)&90.55 &95.07\\
     Bone (Ours)&90.82 &95.55\\
     \hline
     Joint Motion (InfoGCN)&88.60 &93.57\\
     Joint Motion (Ours)&88.85 &94.12\\
     \hline
     Bone Motion (InfoGCN)&87.85 &93.25\\
     Bone Motion (Ours)&88.45 &93.48\\
     \hline
     \bottomrule
   \end{tabular}
  \end{minipage}
  \begin{minipage}{0.49\textwidth}
    \centering
    \caption*{NW-UCLA dataset}
    \begin{tabular}{lll}
      \toprule
      \cmidrule(r){1-2}
      Modality/Method     &Accuracy(\%)  \\
      \hline \hline
      Joint (InfoGCN) &94.40\\
      Joint (Ours) &95.91\\
      \hline
      Bone (InfoGCN) &94.20 \\
      Bone (Ours) &95.26\\
      \hline
      Joint Motion (InfoGCN) &91.59\\
      Joint Motion (Ours) &93.32\\
      \hline
      Bone Motion (InfoGCN) &89.66\\
      Bone Motion (Ours) &92.03\\
      \hline
      \bottomrule
    \end{tabular}
  \end{minipage}
 \end{table}

 \subsubsection*{Alternative TSA MS-SC architecture}
 We compare alternative modeling architecture for TSA MS-SC. Instead of TSA module followed by MS-SC, we utilize MS-SC to generate the
 values $V$ in multi-head self-attention. That is expected to work in two aspects: (i) explicitly learning the short-term temporal motion representation of a joint from its neighboring frames;
 and (ii) introducing beneficial local inductive biases to attentions.
 Therefore, the resulting fused joint representations is expected to have both local temporal relation and global contextual information.
 However, we do not observe significant improvement by this modification only to increase the number of parameters.
 
\section{Conclusion}
In this paper, we propose a novel self-attention and GCN hybrid model for skeleton action recognition.
This framework consists of vanilla self-attentions and GCN of initial physical connections with no need for any hierarchical topology nor sub-graph structures, and nor any data augmentations.
The effectiveness of our method is verified on the benchmark of various datasets.
Our approach is comparable to current state-of-the-arts methods.

\nocite{*}
\bibliographystyle{unsrt}  
\bibliography{references}  
\end{document}